\documentclass[sigconf]{acmart}
\AtBeginDocument{%
 }

\copyrightyear{2024}
\acmYear{2024}
\setcopyright{rightsretained}
\acmConference[CIKM '24]{Proceedings of the 33rd ACM International Conference on Information and Knowledge Management}{October 21--25, 2024}{Boise, ID, USA}
\acmBooktitle{Proceedings of the 33rd ACM International Conference on Information and Knowledge Management (CIKM '24), October 21--25, 2024, Boise, ID, USA}
\acmDOI{10.1145/3627673.3679215}
\acmISBN{979-8-4007-0436-9/24/10}

\makeatletter
\gdef\@copyrightpermission{
  \begin{minipage}{0.3\columnwidth}
   \href{https://creativecommons.org/licenses/by-nc-nd/4.0/}{\includegraphics[width=0.90\textwidth]{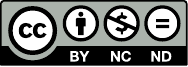}}
  \end{minipage}\hfill
  \begin{minipage}{0.7\columnwidth}
   \href{https://creativecommons.org/licenses/by-nc-nd/4.0/}{This work is licensed under a Creative Commons Attribution-NonCommercial-NoDerivs International 4.0 License.}
  \end{minipage}
  \vspace{5pt}
}
\makeatother

\begin{document}

\title{Preserving Old Memories in Vivid Detail: Human-Interactive Photo Restoration Framework 
}

\author{Seung-Yeon Back$^*$}
\affiliation{%
  \institution{Sungkyunkwan University}
  \city{Suwon}
  \country{Republic of Korea}
}\email{syon1203@g.skku.edu}

\author{Geonho Son$^*$}
\affiliation{%
  \institution{Sungkyunkwan University}
  \city{Suwon}
  \country{Republic of Korea}
}\email{sohn1029@g.skku.edu}

\author{Dahye Jeong}
\affiliation{%
  \institution{Sungkyunkwan University}
  \city{Seoul}
  \country{Republic of Korea}
}\email{gwg03391@g.skku.edu}

\author{Eunil Park}
\affiliation{%
  \institution{Sungkyunkwan University}
  \city{Seoul}
  \country{Republic of Korea}
}\email{eunilpark@skku.edu}

\author{Simon S. Woo}
 \authornote{Corresponding author.} 
\affiliation{%
  \institution{Sungkyunkwan University}
  \city{Suwon}
  \country{Republic of Korea}
}\email{swoo@g.skku.edu}

\renewcommand{\shortauthors}{Seung-Yeon Back, Geonho Son, Dahye Jeong, Eunil Park \& Simon S. Woo}

\begin{abstract}
Photo restoration technology enables preserving visual memories in photographs. However, physical prints are vulnerable to various forms of deterioration, ranging from physical damage to loss of image quality, etc. While restoration by human experts can improve the quality of outcomes, it often comes at a high price in terms of cost and time for restoration. In this work, we present the AI-based photo restoration framework composed of multiple stages, where each stage is tailored to enhance and restore specific types of photo damage, accelerating and automating the photo restoration process. By integrating these techniques into a unified architecture, our framework aims to offer a one-stop solution for restoring old and deteriorated photographs. Furthermore, we present a novel old photo restoration dataset because we lack a publicly available dataset for our evaluation.

\end{abstract}



\maketitle

\def\thefootnote{$^*$}\footnotetext{These authors contributed equally to this work.}\def\thefootnote{\arabic{footnote}}

\vspace{-0.1 em}
\section{Introduction}

The necessity for photo restoration technology arises to restore visual memories or commemorate historical events encapsulated in old photographs. 
In fact, over time, photographs can undergo various forms of degradation, ranging from physical damage, loss of image quality, etc. Also, recently there has been a large demand to colorize gray-scale photos. While restoration and recovery conducted by human experts can ensure quality, it is priced highly, posing accessibility and affordability challenges for many. The resulting cost is recognizably high due to the intense labor and time consumption entailing the whole restoration process such as blemish removal, colorization, resolution enhancement, and texture reconstruction. 
To address the aforementioned challenges from high cost and time, there is a growing interest in developing automated tools by leveraging AI-based methods.

In this work, we propose a modular deep learning framework enabling user-guided restoration for diverse image degradation scenarios, which can provide high quality photo restoration capability.  
We developed the end-to-end photo restoration framework by leveraging Stable Diffusion~\cite{rombach2022high}, GFP-GAN~\cite{wang2021towards}, and DDColor~\cite{kang2023ddcolor} approaches. In particular, we designed and pipelined our restoration frame in the following four stages: 1) major damage removal, 2) noise reduction, 3) facial restoration, and 4) colorization. By integrating them into an unified framework, our approach enables users to tailor the restoration process according to their personalized needs.

In addition, to evaluate the effectiveness of our framework, we create a curated restoration-centric dataset specifically designed for assessing photo restoration algorithms. We conduct a series of comparisons with baseline photo restoration approaches, and demonstrate human evaluations that our methodology effectively restores old photographs. Moreover, by using our framework to restore actual old photos, we show that our approach indeed performs well on real-world data, demonstrated through human evaluation. Finally, we present the importance of user interaction by comparing restoration outcomes generated with and without user input, emphasizing the ability of our framework to incorporate user preferences for a more personalized restoration experience.

\begin{figure*}[tbh]
\centering
\includegraphics[width=0.85\textwidth]{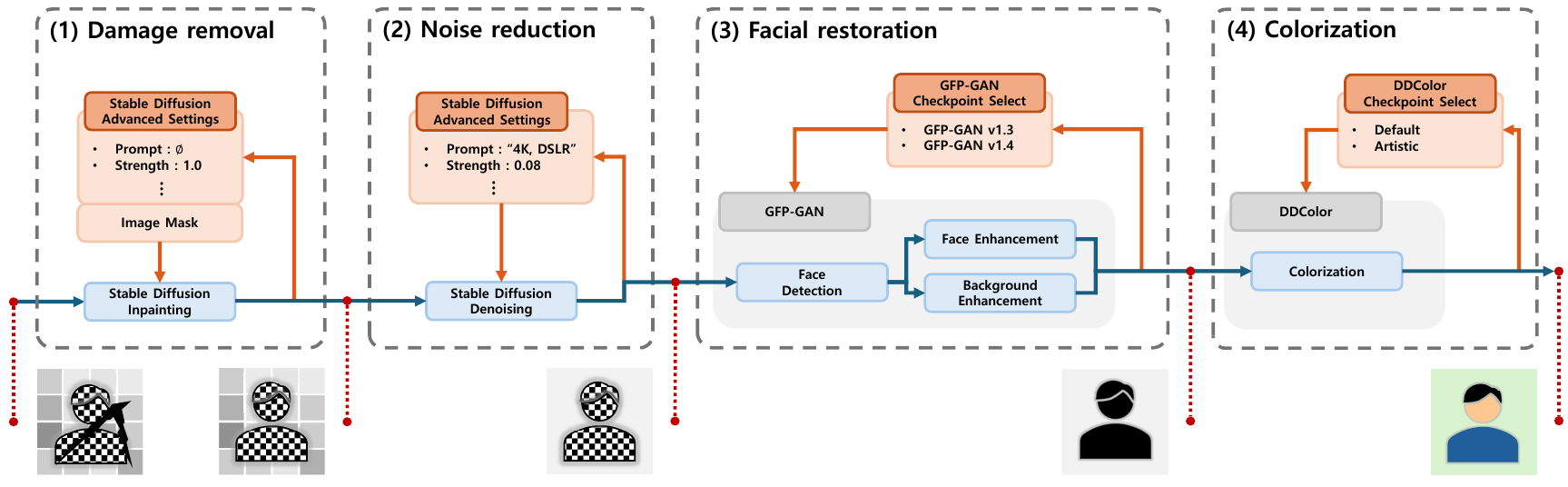}
\vspace{-1 em}
\caption{
Overview of our Restoration Framework showing the process of restoring a given damaged image. The blue lines represent the workflow of the framework, while the orange lines indicate user feedback. The image with red dotted lines connecting different components illustrates the inputs and outputs for each module of the framework.
}
\vspace{-1.5 em}
\label{fig:framework}
\end{figure*}

\vspace{-1 em}
\section{Related Work}

Among several photo restoration models, GFP-GAN~\cite{wang2021gfpgan} provides facial and identity-preserving losses to achieve rapid and effective facial restoration. Also, Restormer~\cite{zamir2022restormer}, SwinIR~\cite{liang2021swinir}, and DDColor~\cite{kang2023ddcolor} are introduced, which are based on transformer architectures to address diverse restoration tasks. In particular,~\cite{wan2020bringing} proposed an approach to tackle the challenge of concurrent image degradations by utilizing multi-domain latent space translation for scratch reduction and facial up-scaling. Recently, diffusion-based restoration models~\cite{kawar2022denoising,zhu2023denoising} 
learn to progressively denoise latent representations, effectively addressing multiple degradations present in old photographs. Moreover, the DDNM~\cite{wang2022zero} allows diffusion model to fill in the null space, enabling it to handle multi-degradation scenarios and perform inpainting and colorization tasks.  Furthermore, DDNM demonstrates inherent robustness against real-world noise. Recent research has also explored the application of Latent Diffusion Models (LDMs) for photo restoration~\cite{rombach2022high, lin2024diffbir}. The generative power of LDMs excels at inpainting and super-resolution, particularly for high-resolution images. 

\vspace{-1 em}
\section{Architecture}

Our framework consists of four stages: 1) major damage removal, 2) noise reduction, 3) facial restoration, and 4) colorization. Each stage is designed to be visually inspected and adjusted based on user feedback to achieve the most pleasant and natural outcomes for users. By incorporating user feedback, our framework provides flexibility, enabling users to achieve their desired restoration outcomes for each individual photo. We employed state-of-the-art restoration methodologies within our framework to incorporate their distinctive strengths. Effectively, these methodologies influence each other based on the order of the stages. Therefore, we followed the sequence of severe damage repair, noise reduction, facial restoration, and colorization. We explain more details for each stage as follows:

\noindent \textbf{1) Damage Removal.} Recently. Stable Diffusion (SD)~\cite{rombach2022high}, a type of LDM, was utilized for in-painting due to its zero-shot capability. To maximize the focus on the sole restoration of the image, we utilize SD for a text-unconditional synthesis. We first aim to tackle the major defects using SD in-painting process.

\noindent \textbf{2) Noise Reduction.} While SD~\cite{rombach2022high} was originally designed to denoise procedurally added Gaussian noise, it is not designed to handle real-world noised images. However, due to its general applicability, it can be employed for denoising tasks as well.
We find that removing noise early in the process is more effective for facial restoration and colorization at the later stage by providing a cleaner image for these more detailed stages. The assumption is that the expected clean image is in fact noisy, and by treating it as though \( t \) has not fully converged to 0, the input image is processed to converge to timestep 0, effectively reducing the noise level. We set the default prompt to ``4K, DSLR" and the strength to 0.08, using DDIM steps proportional to the noise intensity to which we wish to remove.

\noindent \textbf{3) Facial Restoration.} We use GFP-GAN~\cite{wang2021towards} for facial restoration, where GFP-GAN can detect and restore faces while maintaining the original identity. And, GFP-GAN can also effectively enhance the background information through super-resolution~\cite{wang2021realesrgan}, thereby providing comprehensive image restoration. Likewise, restoring facial features before colorization can ensure that critical identity-preserving details are accurately recovered, which is crucial for the visual integrity of the restored image and establishes a foundation for the colorization process.

\noindent \textbf{4) Colorization.} We integrated DDColor~\cite{kang2023ddcolor}, which offers high resolution and generality, and effectively colorizing diverse backgrounds and artifacts. Applying colorization at the last stage can guarantee that all preceding structural and detail restorations are completed, producing a more cohesive and natural-looking final result. Moreover, as described in Figure~\ref{fig:framework}, we enabled users to modify settings at each stage. And, users can possibly adjust parameters from the default settings based on their preferences, such as the degree of noise, the presence of damages, and the desired color.

\begin{figure}[tbh]
\centering
\includegraphics[width=0.97\columnwidth]{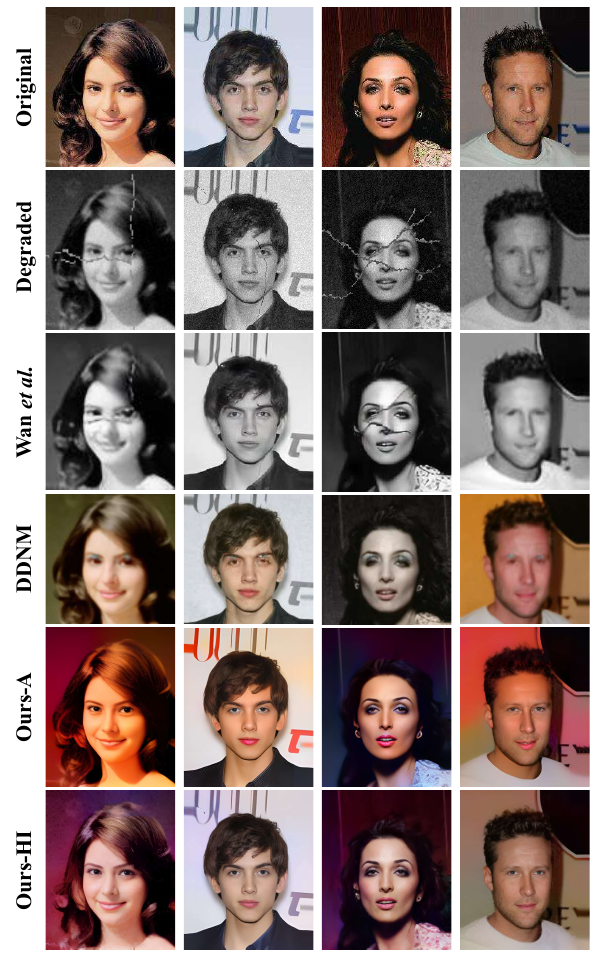}
\vspace{-1.5 em}
\caption{
Comparison with baseline restoration frameworks. Ours-A represents the results when our pipeline is utilized automatically with the given mask, while Ours-HI denotes the results when our pipeline is manually used with human interaction. 
}
\vspace{-2 em}
\label{fig:qualitative}
\end{figure}

\vspace{-1.5 em}
\section{Dataset}
\textbf{Old Photo Dataset. }Currently, there is no publicly available datasets for old photo restoration. Furthermore, the absence of ground-truth masks for rips and major damage areas in facial images poses a significant issue for automated in-painting task. Therefore, a dataset that can closely mimic old photos is required for our task. 
To construct the new old photo dataset, we randomly select 5,000 images each from UTKFace~\cite{zhang2017age} and CelebA~\cite{liu2015faceattributes}. Then, we apply four different modification methods to provide the noise, blur, deteriorating, and color fading effects commonly observed in real-world old photographs. First, we convert the original images to gray-scale to render them in black and white. This step replicates the monochromatic look of early photography. Next, we randomly apply various blurring methods, including Gaussian, median, and bilateral filters, combined with down-scaling techniques at different scales to degrade the image quality. This process simulates the lower resolution and blurry appearance of old worn-out photographs. Next, we generate images and masks with various physical damage patterns using a crack generator~\footnote{\url{https://github.com/YoonSungLee/crack_generator}}. This technique adds realistic damage patterns to the image, mimicking the physical deterioration often seen in aged photos. Lastly, we apply Gaussian noise with random scaling factors to the images. This step adds a layer of randomness and imperfections, further producing degraded images similar to the real world old photos. Through a sequence of modifications, we generate four types of images at each stage:

\begin{itemize}
    \item Gray-scale 
    \item Gray-scale + Blurring or Down-scaling
    \item Gray-scale + Blurring or Down-scaling + Crack
    \item Gray-scale + Blurring or Down-scaling + Crack + Noise
\end{itemize}

Examples of this dataset can be seen in the degraded section of Figure~\ref{fig:qualitative}. The `original' represent the unaltered data, while the `degraded' are generated using our methodology.

\begin{figure}[t]
\centering
\includegraphics[width=0.95\columnwidth]{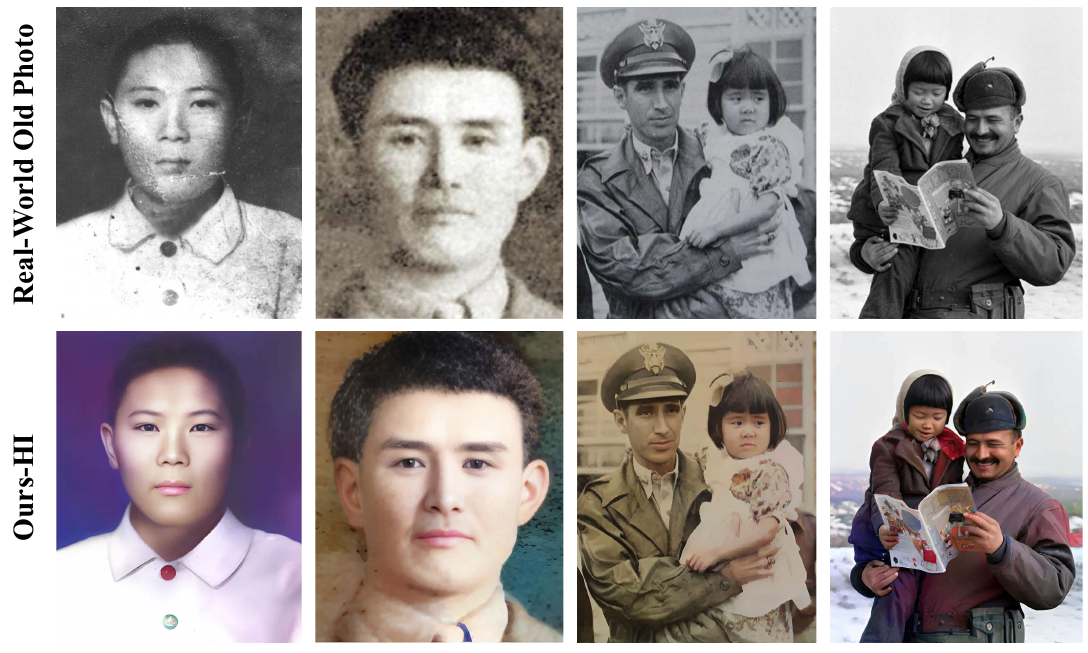}
\vspace{-1.5 em}
\caption{
Additional results of restoring real-world old photos using our framework with human interaction.
}
\vspace{-3 em}
\label{fig:test}
\end{figure}

\noindent \textbf{Real-World Old Photo Dataset. }We used old photographs of soldiers and wartime scenes from Korean War, provided by the Ministry of Patriots and Veterans Affairs in South Korea~\cite{website:minister}. Given that this dataset are actual data from old photographs, it presents a complex combination of image degradation. Examples of this dataset are presented in Figure~\ref{fig:test}. We use this dataset for evaluating our framework.

\vspace{-0.5 em}
\section{Experiments}
\textbf{Experimental Setting. } We configured the evaluation process as a one-stop end-to-end framework for all models. This enables us to compare user preferences in an automated manner, without requiring user feedback, similar to other prior methods. For human-guided damage restoration masks, we excluded framework proposed by Wan \textit{et al.}~\cite{wan2020bringing}, which automatically generates damage mask. For the models which requires hand-crafted damage masks, we used masks by padding the damaged regions used during synthesis to fully cover the damage area. 
For damage removal, we utilized a null text prompt and set the strength 1.0, DDIM ~\cite{song2020denoising} step 30 and CFG ~\cite{ho2022classifier} 1.0. For noise reduction, the prompt was ``4K, DSLR" with a strength 0.008, DDIM step 50 and CFG 3.0. For face restoration, we utilized version 1.3 of GFP-GAN ~\cite{wang2021gfpgan}. For colorization, we employed the checkpoint ``modelscope" in DDColor~\cite{kang2023ddcolor}.

\noindent \textbf{Qualitative Comparison. }The results from qualitative comparison are described in Figure~\ref{fig:qualitative}, where our dataset encompasses various degradation while ensuring that the facial identity remains intact overall. When restored using the framework proposed by Wan \textit{et al.}~\cite{wan2020bringing}, the limitations in damaged area detection resulted in incomplete removal of major damage. Additionally, the absence of a colorization module prevented the introduction of color to the images. In the case of DDNM~\cite{wang2022zero}, while damage were effectively removed through in-painting given ground-truth mask, certain images exhibited incorrect restoration of the eye region, as the model was trained to fit a specific dataset. Furthermore, the low resolution of the resulting images led to a perceived reduction in image quality. In contrast, our automatically restored images using the default setting are free from issues related to damage removal, resolution, and generalizability. One potential drawback is that the color may appear overly saturated for certain images. This is attributable to the fact that the colorization module, DDColor~\cite{kang2023ddcolor}, provides checkpoints trained on ImageNet~\cite{russakovsky2015imagenet}, leading to a mismatch with the test distribution. Through human interaction, by selecting appropriate values and checkpoints for each image, outstanding results can be obtained.

\begin{figure}[tbh]
\centering
\vspace{-0.5 em}
\includegraphics[width=0.9\columnwidth]{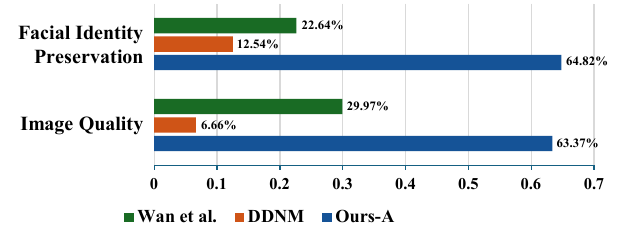}
\vspace{-1 em}
\caption{
Comparison of human evaluation ratios across two questions.
}
\vspace{-1.2 em}
\label{fig:human_eval}
\end{figure}

\noindent \textbf{Human Evaluation. }Our goal in photo the proposed restoration framework is to achieve high-resolution, noise-free, and colored images. Given this objective, commonly used datasets with varying image quality can yield metric evaluations that differ from human perceptions. Therefore, we conducted human evaluation to identify the most visually natural and satisfactory restoration methods. We created a survey with two questions on Google Forms to collect preferences of restored images for Wan~\textit{et al.}~\cite{wan2020bringing}, DDNM~\cite{wang2022zero}, and our framework. The first question presented a degraded image, and participants were asked to choose the well-restored image. In the second question, participants were shown the original image and asked to select the image with the most similar facial identity.
In this survey, we obtained responses from each participant for 15 questions per question type, totaling 30 questions. Participants were asked to choose their preferred results from the three different methodologies for each question. With 101 participants, we received 3,030 evaluations. As shown in Figure ~\ref{fig:human_eval}, our framework received 63.37\% and 64.82\% support for the respective questions, indicating a higher human preference in terms of quality and identity preservation in our method compared to other approaches. Through the human evaluation, we can clearly verify that the overall quality of restoration achieved with our method surpasses existing approaches. Additionally, our methodology successfully restores severely damaged images, while preserving their original features.
\begin{figure}[tbh]
\centering
\vspace{-0.3 em}
\includegraphics[width=0.95\columnwidth]{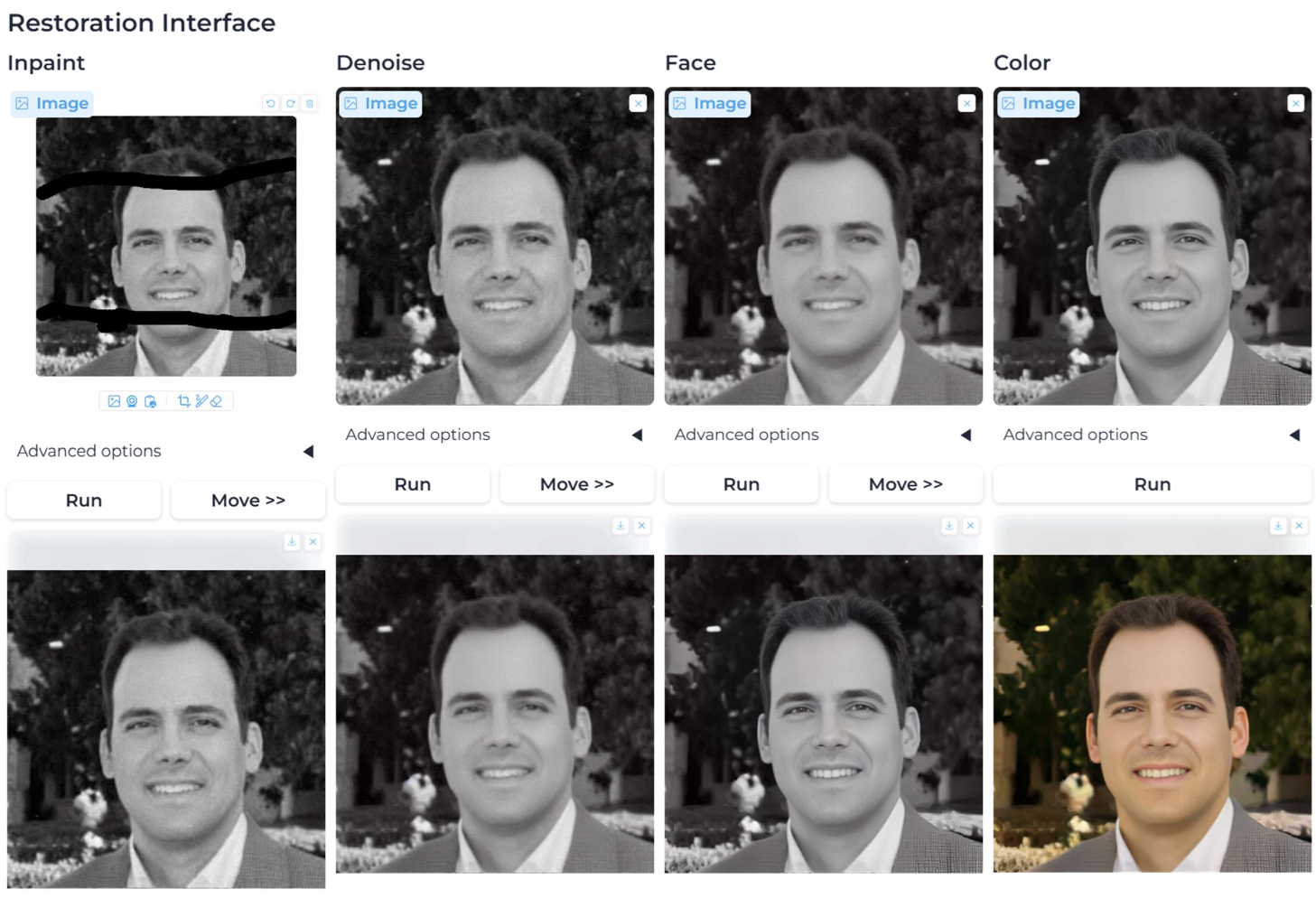}
\vspace{-1 em}
\caption{
Restoration interface implementing our proposed framework.
}
\vspace{-1.5 em}
\label{fig:gui}
\end{figure}

\section{Tool Usage}

Our interface is depicted in Figure~\ref{fig:gui}. We utilized Gradio~\cite{abid2019gradio} to build an intuitive interface for users to interact with and customize the process. In this context, `user' refers to anyone utilizing our GUI to restore photos. Our interface was designed so that users could view the original image at the top and the restored version below. Users can proceed to the next restoration stage by clicking the `Move' button once satisfied with the current outcome. This iterative process allows users to continuously refine the restoration until the desired result was achieved.
Users are allowed to adjust parameters 
and seeds to repair tears and major damages through in-painting at the first stage. Drawing tools are available to select the damaged area, enabling users to provide a mask for restoration. Similar adjustments are available to reduce noise, refining the image quality further at the next stage. Face restoration stage focuses on facial enhancement, allowing users to apply face-specific restoration and background super-resolution selecting two model weights. For the final step, users can select the model weight to achieve the desired color tones. By structuring the user interface in this manner, we ensure that each restoration stage is meticulously handled, allowing users to achieve a high level of customization and satisfaction with the final restored image. Our demonstration process can be viewed by referring to the video~\footnote{\url{https://youtu.be/dbZFPJ7Or3I}}.

\section{Conclusion}

In this paper, we presented an intuitive human interactive restoration framework from real-world old photos. We also created a dataset containing pairs of original and degraded images, allowing for effective qualitative experiments by enabling the evaluation of restoration quality. In addition, human evaluation effectively verifies whether the restoration addresses practical challenges, which can be more accurately assessed through direct feedback from human participants. Through our experiment, we demonstrate the effectiveness of our proposed framework, operating at a level satisfactory to the general public. We hope that this framework can be widely adopted for the restoration of various images to preserve personal memories as well as to commemorate monumental events from old, and severely damaged photos.

\begin{acks}

We thank Jaseng Hospital of Korean Medicine and Ministry of Patriots and Veterans Affairs in Republic of Korea for sponsoring this research and providing sample images. Also, this work was partly supported by Institute for Information \& communication Technology Planning \& evaluation (IITP) grants funded by the Korean government MSIT: (RS-2022-II221199, Graduate School of Convergence Security at Sungkyunkwan University), (RS-2024-00337703, Development of satellite security vulnerability detection techniques using AI and specification-based automation tools), (RS-2022-II221045, Self-directed Multi-Modal Intelligence for solving unknown, open domain problems), (RS-2022-II220688, AI Platform to Fully Adapt and Reflect Privacy-Policy Changes), (RS-2021-II212068, Artificial Intelligence Innovation Hub), (RS-2019-II190421, AI Graduate School Support Program at Sungkyunkwan University), (RS-2023-00230337, Advanced and Proactive AI Platform Research and Development Against Malicious Deepfakes), and (RS-2024-00356293, AI-Generated Fake Multimedia Detection, Erasing, and Machine Unlearning Research).

\end{acks}

\balance

\bibliographystyle{ACM-Reference-Format}
\bibliography{acmart}

\appendix

\end{document}